\documentclass[runningheads]{llncs}
\usepackage[T1]{fontenc}
\usepackage{graphicx}
\usepackage{adjustbox}
\usepackage{booktabs}
\usepackage{multirow}
\usepackage{pifont}
\usepackage{hyperref}
\usepackage{listings}
\usepackage{xcolor}
\usepackage[inline]{enumitem}
\usepackage[colorinlistoftodos,prependcaption,textsize=small]{todonotes}
\usepackage[ruled,vlined]{algorithm2e}

\begin{document}
\title{Explaining Reinforcement Learning Decisions in Self-adaptive Systems}
%\author{annon}
\author{Jasmina Gajcin \and
Juan C. Rosero \and
Ivana Dusparic%\inst{1}
}
%
% \authorrunning{Gajcin et al.}
%\institute{annon}
\institute{School of Computer Science and Statistics \\
Trinity College Dublin \\
\email{\{gajcinj,roserolj,duspari\}@tcd.ie}}

\maketitle          

\begin{abstract}
Reinforcement Learning (RL) has been extensively used in autonomous and self-* systems, but RL policies, especially deep RL ones relying on neural networks, lack transparency and are difficult to understand. This can lead to diminished user trust, and makes for a more challenging verification of systems. To address this challenge, this paper introduces Explanations using Alternative Realities for Reinforcement Learning (EARL), a Python library to produce counterfactual explanations in RL settings. This library allows the user to produce explanations by exploring What-if scenarios to clarify agent behavior by comparing possible outcomes. Counterfactual explanations have been shown to be intuitive and user-friendly in psychology research, but have only recently been explored in RL, with existing implementations usually limited to toy examples and benchmarks. EARL supports counterfactual explanation generation in realistic RL-based self-adaptive systems. To demonstrate its applicability, we demonstrate its use in a simulation of CitiBikes \cite{MARO_MSRA}, a self-adaptive bike-sharing system, and we provide evaluations showing how it performs in real applications.

\end{abstract}
\section{Introduction}
\label{intro}

Self-adaptive systems dynamically modify their behavior in response to environmental changes \cite{macias2013self}. RL is one of the most widely used machine learning (ML) techniques for enabling adaptations \cite{saputri2020application,d2019learning}, allowing agents to learn optimal policies through interaction and feedback from the environment \cite{Sutton1998}. However, policies learned through RL, especially those using neural networks, are often opaque and difficult to interpret, due to the state-action mapping process inherent to neural networks.

As self-adaptive systems are expected to operate alongside and assist users, including non-experts, there is a growing need for explanations to foster trust, facilitate verification, and support collaboration \cite{van2024towards,yigitbas2021enhancing,li2020explanations}. The field of explainable reinforcement learning (XRL) addresses this need, but many of its methods are designed for developers, relying on low-level visualizations or global approximations of black-box models \cite{skirzynski2021automatic,puri2019explain}, and usually requiring a high level of technical expertise. 

In contrast, counterfactual explanations offer high-level, user-friendly insights by showing how outcomes would change under different conditions. Counterfactual explanations explain a decision by showing "What if" scenarios, illustrating why the agent's decision was chosen. For example, an agent controlling a walking robot might justify turning to the left by showing that if it had kept going straight, it would have fallen into a hole. These have been widely adopted in supervised learning \cite{wachter2017counterfactual,verma2021counterfactual}, but remain underexplored in RL \cite{gajcin2024redefining}. Crucially, current efforts are hindered by the lack of openly available tools for generating and evaluating counterfactual explanations in realistic RL settings.

To address this gap, we introduce EARL, a Python library for generating counterfactual explanations in RL. This library supports multiple explanation methods and enables users to explore "what-if" scenarios in complex environments. Unlike existing implementations, which are typically tied to a single explanation method or restricted to benchmark environments, EARL provides a unified and extensible framework that integrates multiple state-of-the-art counterfactual explanation methods under a common interface. The library includes reusable model wrappers, a standardized evaluation framework, and reference implementations that facilitate reproducible experimentation across different RL agents and environments. Additionally, EARL extends existing methods enabling their application to structured state representations common in self-adaptive systems. Our artifact is designed to be easily extendable, reproducible, and applicable to realistic tasks beyond toy domains. We demonstrate its use in CitiBikes, a simulated self-adaptive bike repositioning system inspired by New York City's bike-sharing program \cite{MARO_MSRA}. We evaluate four counterfactual explanation methods, showing how they can reveal the reasoning behind RL agent behavior in a practical scenario. The artifact includes implementations of the explanation methods, example usage, and experimental setup, and is available on github \footnote{\href{https://github.com/JuanK120/EARL}{https://github.com/JuanK120/EARL}}.

The rest of the paper is structured as follows. In section \ref{SectionRelatedWork} we give background on RL, XRL and counterfactual explanations and their usage on autonomous systems. Next, in section \ref{sectionGenerationMethods} we give a general overview of the methods our library contains. Section \ref{sectionLibraryStructure} explains the  composition and structure of our library. Finally, section \ref{SectionCitiBikes} demonstrates our example application of EARL in a real-world scenario.

\section{Related Work}
\label{SectionRelatedWork}

In this Section, we first provide an overview of RL applications in self-adaptive systems (Section \ref{SectionRelatedWork:RLforSAS}). Section \ref{SectionRelatedWork:XRL} covers the topic of explainable RL, while counterfactual explanations are presented in Section \ref{SectionRelatedWork:CFSF}. 

\subsection{RL for Self-Adaptive Systems}
\label{SectionRelatedWork:RLforSAS}

RL is a powerful ML approach for developing complex, multi-objective policies. RL agents adapt to uncertain conditions, changing their behavior depending on the environment. This makes RL suitable for developing self-adaptive systems, with some of the applications including data center cooling \cite{heimerson2022adaptiveCooling}, drone navigation \cite{hossain2023covernav}, and self-adaptive servers \cite{rosero2024DwnEws}. 
However, RL decisions are often difficult to interpret, affecting their applicability to high-risk real-life problems \cite{milani2024explainable}.  

\subsection{Explainable RL}
\label{SectionRelatedWork:XRL}

RL agents, particularly those using neural networks, often make decisions that are difficult to interpret \cite{milani2024explainable}. Explainability is important for multiple reasons: developers need it for verification and debugging \cite{gajcin2022reccover,metzger2023user}, experts are less likely to trust opaque systems \cite{cadario2021understanding}, and legal frameworks like the EU’s GDPR may require explanations for fairness \cite{sovrano2018making}.
XRL focuses on methods to explain agent decisions \cite{milani2024explainable}, and they are categorized in global (explaining overall behavior) or local (explaining individual decisions) methods. For example, behavior summaries help users understand agent strengths \cite{amir2018highlights}, while saliency maps show which input regions influenced a specific action \cite{greydanus2018visualizing}. However, most XRL methods target developers and offer low-level, task-specific insights. Non-expert users instead need high-level, actionable explanations to effectively interact with RL systems.

\subsection{Counterfactual Explanations}
\label{SectionRelatedWork:CFSF}

The objective of counterfactual explanations is to explain why an outcome occurred by presenting an alternative scenario with a different result. This contrastive nature aligns with human reason, making them intuitively appealing \cite{miller2019explanation}, and their causal framing helps users assign blame \cite{byrne2019counterfactuals}. These qualities make counterfactuals particularly user-friendly. While extensively studied in supervised learning \cite{verma2021counterfactual}, counterfactual methods in RL remain limited \cite{olson2019counterfactual,huber2023ganterfactual,gajcin2024redefining}. In RL, prior work has also shown that counterfactual explanations can improve users' understanding of agent behavior and support decision-making in controlled user studies \cite{Gajcin2023RACCERTR}, such methods aim to explain why an agent selects action $a$ in state $s$ by generating a similar state $x'$ where a different action $a'$ would be taken.
However, despite their promise, counterfactuals in RL have so far been applied only in simplified environments like Gridworld and Atari games.

\section{Counterfactual Generation Methods in RL}
\label{sectionGenerationMethods}

In this section, we provide an overview of counterfactual generation methods in RL implemented in our artifact.

\subsection{GANterfactual-RL} 
\label{sectionGenerationMethods:Ganterfactual}

GANterfactual-RL \cite{huber2023ganterfactual} is a dataset-based counterfactual generation method for RL. It requires only a dataset of agent transitions but as such does not consider the sequential or stochastic nature of the environment. It frames counterfactual generation as a domain transfer problem, splitting the state space into domains based on the agent's action and translating a state from one domain to another. Based on the StarGAN architecture, it uses a generator to produce realistic states and a discriminator to distinguish generated from real states. The original GANterfactual-RL implementation (available with the original publication) was aimed only at image-based inputs; in our library we implement and provide a non-image based version to enable its wider applicability. 

\subsection{RACCER} 
\label{sectionGenerationMethods:RACCER}

RACCER (\underline{R}eachable and \underline{C}ertain \underline{C}ounterfactual \underline{E}xplanations for \underline{R}L) \cite{gajcin2023raccer} is an RL-specific approach that searches for counterfactuals in an agent's execution by exploring the states that can be reached from the original state, and therefore requires access to the agent's execution environment. 
Three optimization objectives are used to evaluate and search for counterfactuals. Firstly, RACCER aims to minimize reachability, measuring the distance between the original state $x$ and counterfactual $x'$ as the number of RL actions $A$ between them. Next objective is fidelity, which ensures that the counterfactual $x'$ is reached from the original state via a path likely under the policy being explained, ensuring the counterfactual is representative of the agent's policy. Finally, RACCER ensures that counterfactuals are reached with high certainty even in stochastic environments, this is done by optimizing stochastic uncertainty. 

There are three variants of RACCER, each varies in the way they search and generate counterfactuals: 

\subsubsection{RACCER-HTS}
\label{sectionGenerationMethods:RACCERHTS}
In this variant, we use a heuristic tree search (HTS) to find suitable counterfactuals. RACCER combines reachability, fidelity, and stochastic uncertainty into a loss function to evaluate action sequences, balancing exploration and exploitation. It then filters results to ensure the counterfactual state $x'$ leads to the target action $a'$.

\subsubsection{RACCER-Advance}
\label{sectionGenerationMethods:RACCERADVANCE}
This variant reduces computational cost by replacing tree search with an evolutionary algorithm (NSGA-II). RACCER-Advance generates counterfactuals starting from the agent’s current state $x_n$, exploring how future actions could lead to an alternate decision. Formally, given a state $x_n$, it searches for a sequence of actions $A$ such that applying $A$ from $x_n$ leads to a counterfactual state $x'$. This variant allows users to ask questions like: \textit{“What could change in the future for a different decision?”} It retains the three original objectives and optimizes them through NSGA-II, producing diverse and representative counterfactuals.

\subsubsection{RACCER-Rewind}
\label{sectionGenerationMethods:RACCERREWIND}
RACCER-Rewind generates counterfactuals by modifying the agent’s past. It takes as input the past state-action sequence $(x_{n-k}, a_{n-k}),$ $ \dots, $ $ (x_{n-1}, a_{n-1})$ leading to the current state $x_n$ and explores alternative past sequences that could have resulted in a different outcome. It searches for a modified action sequence $A$ starting from $x_{n-k}$ that results in a counterfactual $x'$ leading to a different decision. This variant answers questions like: \textit{“What needed to be different in the past?”} Like RACCER-Advance, it employs NSGA-II to optimize the same three objectives.

%Both RACCER and Ganterfactual-RL have been tested on Stochastic Gridworld and Frozenlake \cite{gajcin2023raccer}

\section{Library Structure}
\label{sectionLibraryStructure}

In this section, we introduce EARL's structure, and main features. The library is designed in a modular structure and is extensible to multiple models and implementations. Our library's main components are: 
\begin{enumerate*}
    \item the explanation methods,
    \item model wrappers, and
    \item the evaluation interface.
\end{enumerate*}
Additionally, we include basic implementations of both a simple Deep Q-learning algorithm, and a PPO Algorithm. 

The \textbf{Explanation Method} module contains the core logic for generating counterfactual explanations. Each explanation algorithm is implemented as an independent module that exposes a common interface, allowing users to select the desired strategy without modifying the rest of the library. The currently supported methods are GANterfactual, RACCER-HTS, RACCER-Rewind, and RACCER-Advance, described in Section \ref{sectionGenerationMethods}.

The \textbf{Model wrapper} module provide a common interface between EARL and RL agents. Thee are interfaces created to facilitate implementation on different RL models so EARL is able to properly interact with them. These models assume the existence of either a pre-trained model, or the usage of the default agent methods provided in EARL. These wrappers add a common set of functions to the RL agents that the explanation methods can call to retrieve the agent’s chosen actions and related decision metrics. Consequently, explanation algorithms can be applied to different RL models without modification.

The \textbf{Evaluation interface} module provides reusable tools for comparing explanation methods. EARL provides two evaluation mechanisms for assessing quality and usefulness of the methods: metric-based evaluation and runtime logging. Metric-based evaluation is provided on the evaluation modules of each implementation, they compute metrics such as coverage, generation time, similarity, plausibility, and diversity. These metrics provide a comparison for the explanation methods and help assess their performance. Users can also extend these metrics to define evaluation criteria that better suit their domains. Runtime logging captures agent actions, explanation outputs, and environment responses in real time during simulation.

The \textbf{Learning algorithm} module provides a basic implementation for DQN and PPO algorithms using Stable Baselines3 framework. These implementations are included as default options within their respective model wrappers, allowing users to immediately test explanation methods without requiring a custom model. When no pre-trained model is provided, the wrapper will automatically train a new model using standard hyperparameters in a specified environment. This simplifies the experimentation process and ensures compatibility with the explanation and evaluation components of the library.

\begin{figure}[ht]
    \centering
    \includegraphics[width=0.8\linewidth]{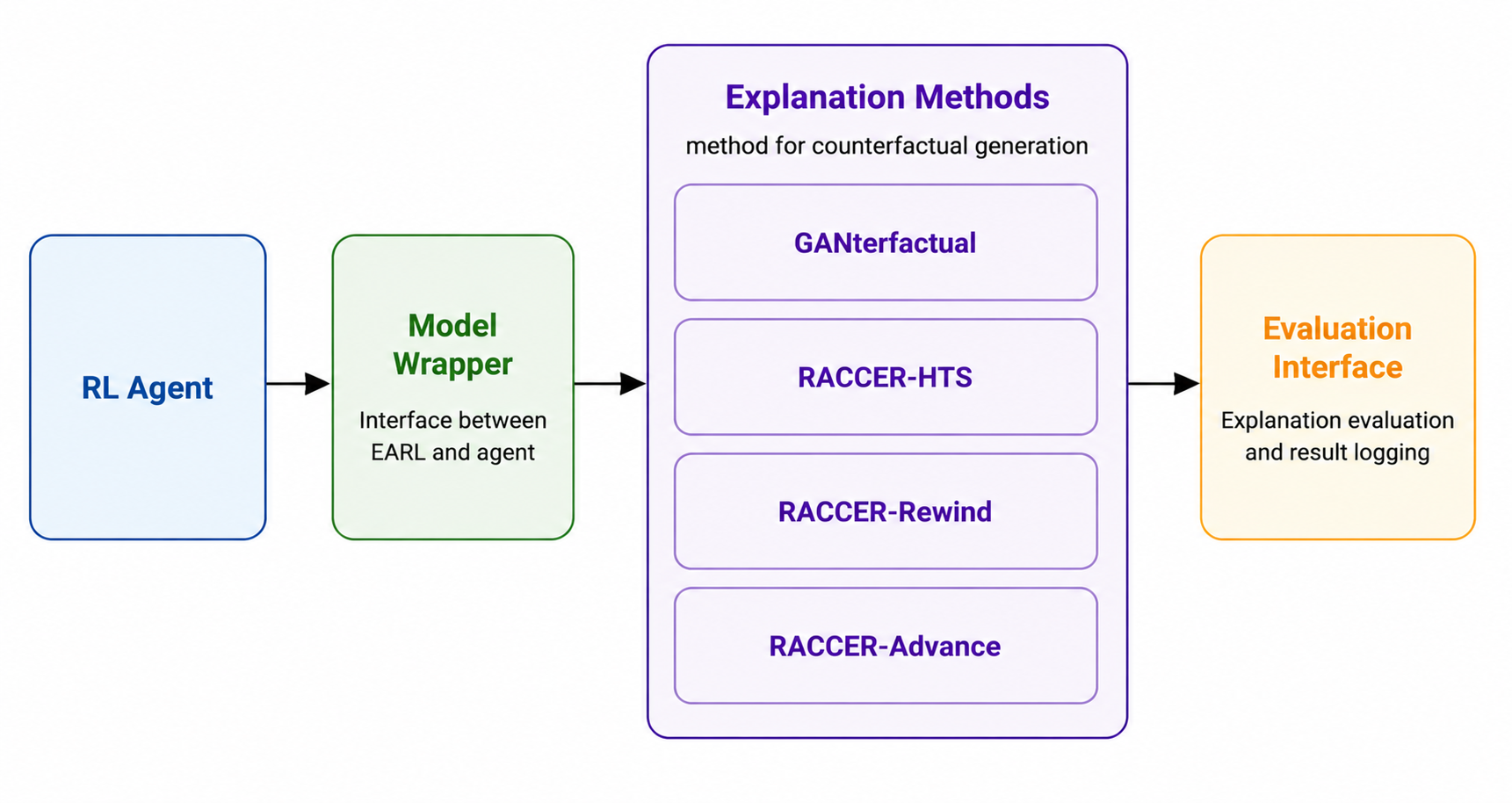}
    \caption{Workflow of EARL library} 
    \label{FigureEARLDiagram}
\end{figure}

Figure \ref{FigureEARLDiagram} shows the overall interaction between the components is as follows: the model wrappers provide a uniform interface to the RL agents, then the explanation methods query these wrappers to generate the counterfactual explanations, and the evaluation interface assesses the generated explanations independently of the underlying RL algorithm.

\section{Utilizing EARL for Explainable RL-based Bike Sharing System}
\label{SectionCitiBikes}

In this section, we introduce an implementation example and describe how RL can be applied to a real-life bike sharing environment, CitiBikes. Section \ref{SectionCitiBikes:envDescription} describes the RL environment for bike sharing, section \ref{SectionCitiBikes:PPO} details the training of a black-box policy for this task, and section \ref{SectionCitiBikes:EARLIntegration} shows the integration process with EARL. 

\subsection{CitiBikes Environment}
\label{SectionCitiBikes:envDescription}

CitiBikes \cite{MARO_MSRA} is an RL environment modeled after New York City's bike-share system, featuring a network of docking stations where bikes are rented and returned. Demand varies with location, time, and weather, often causing imbalances: some stations may be overloaded while others face shortages. To address this, bike repositioning is required to meet dynamic, stochastic demand. We simulate a topology of five stations (S1–S5). Station S1 is balanced, S2 and S5 receive more returns than rentals, and S3 and S4 frequently face shortages due to high demand.

The state space includes 38 features capturing system-wide and per-station data such as bike count, capacity, climate, day type (e.g., weekend, holiday), fulfillment rate, shortages, failed returns, and trip requirements. The action space is multi-discrete: the agent chooses a source station, a destination, and the number of bikes to transfer (up to 10). At each step, the agent observes station data and aims to minimize shortages. Rewards penalize both shortages and bike transfers:

\begin{equation}
R(s, a, s') = -100 \cdot bike_shortage - 0.01 \cdot n
\end{equation}

\subsection{Black-box Policy}
\label{SectionCitiBikes:PPO}

\begin{figure}[ht]
    \centering
    \includegraphics[width=0.8\linewidth]{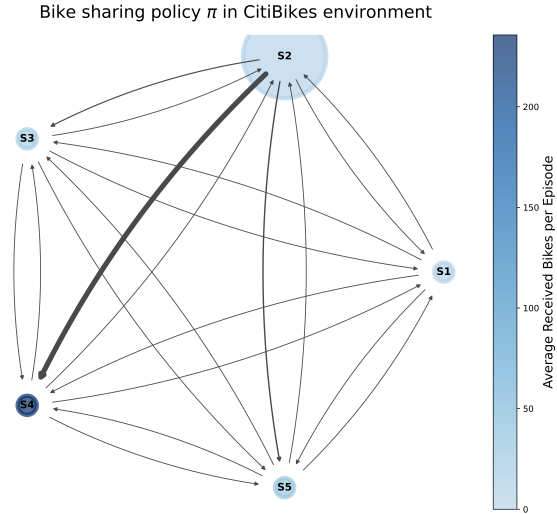}
    \caption{CitiBikes Repositioning Policy $\pi$: Over 100 episodes, bikes are moved among five stations. Node size shows sent bikes; color shows received bikes. Station $S4$ receives the most, mainly from $S2$.} 
    \label{FigureTrainedPolicy}
\end{figure}

To generate explanations for CitiBikes, we start by training a black-box policy $\pi$. We use the PPO algorithm \cite{schulman2017proximal}, but the methods covered in this work are completely model-agnostic and can be used to explain the behavior of any RL policy. Note that our goal here is only to obtain a sensible policy that can be explained by counterfactual methods, rather than to achieve state-of-the-art performance on CitiBikes tasks. Figure \ref{FigureTrainedPolicy} illustrates the bike transfer between stations managed by policy $\pi$ over $100$ episodes, Intuitively sending bikes from the nodes recieving more returned bikes, stations s2 and s5, to the higher demand stations s3 and s4.
%The training parameters and performance of the black-box policy $\pi$ are given in Table \ref{table:BBTraining}. 

\subsection{Integrating EARL to the PPO Policy}
\label{SectionCitiBikes:EARLIntegration} 

Once the PPO policy is trained, integration of the model and EARL to generate the counterfactuals is done as follows: 

\subsubsection{Environment Modifications}
\label{SectionCitiBikes:envModifications}

To make sure the same input format is used at all times, we modified the CitiBikes environment to apply the same preprocessing pipeline during both training and explanation. In particular, we implemented a custom \textit{generate\_obs} method, which creates a consistent state representation using station, decision, and shared features. This method is called both at the end of each environment step and during state resets, ensuring a unified representation throughout all learning and explanation stages.

Additionally, we enforced the action structure required by EARL, using a MultiDiscrete action space with three dimensions: source station, target station, bike count. This is set in the environment's \textit{\_\_init\_\_} method:
\begin{lstlisting}[language=Python]
self.action_space = MultiDiscrete([self.num_stations, self.num_stations, self.max_bike_transfer])
\end{lstlisting}

We also needed a way to reset the environment to arbitrary intermediate states. This is crucial for counterfactual generation, where the agent must try different actions from the same prior state. We addressed this by adding a \textit{set\_nonstoch\_state} method. Below is an algorithm giving an overview of the method:

\section{Evaluation}
\label{SectionEvaluation:EvaluationSetup} 
 
To evaluate EARL in CitiBikes, we select $100$ informative states (out of 306 total states encountered during execution) and generate counterfactual explanations using the four implemented methods. We select those states where there is a large difference between choosing the best and the worst action, inspired by the HIGHLIGHTS algorithm \cite{amir2018highlights}. %As we deal with multi-discrete actions, we choose states where there is a large difference in Q-values along at least one action dimension. 

Our aim is to offer insight into the question: ``Why was action $a$ chosen in state $x$?'' by offering a counterfactual state $x'$ where a different action would have been chosen. Due to the large action space, in this example we do not fix a specific target action but consider all counterfactuals that change the action $a$. The parameters used are summarized in Table \ref{table:TrainingParamsMethods}. 

\subsection{Evaluation Metrics}
\label{SectionEvaluation:EvaluationMetrics} 

We use an evaluation framework composed of  metrics \cite{gajcin2024redefining,gajcin2023raccer} centered around 4 different key parameters: scalability, feature similarity, plausibility, and diversity. These parameters are evaluated as follows:

\begin{enumerate}
    \item Scalability: Refers to how well the approaches scale in high-dimensional tasks. Evaluated in three metrics: \begin{enumerate*}
        \item coverage: percentage of input facts for which an explanation was successfully generated, 
        \item pre-training time, and 
        \item average counterfactual generation time.
    \end{enumerate*} 
    
    \item Feature-based Similarity: two metrics are used: proximity and sparsity. We use definitions of proximity and sparsity used to evaluate the GANterfactual-RL \cite{huber2023ganterfactual} and RACCER \cite{gajcin2023raccer} approaches.
    
    \item Plausibility: Refers to the generation of realistic instances. Plausibility depends on the specific task and the question that the explanation is answering. For the CitiBikes tasks we define three constraints that need to be satisfied: \begin{enumerate*}
        \item A bike station cannot store more bikes than its capacity, 
        \item A bike station cannot loan out more bike than its capacity, and
        \item The shortage of bikes at a station has to be lower than its capacity.
    \end{enumerate*}
    
    \item Diversity: Addresses the need to generate multiple counterfactuals reflecting different factors causing a certain outcome. We evaluate diversity by measuring the number of generated counterfactuals per factual query and pair-wise feature distance within the solution set. 
\end{enumerate}

\begin{table}[t]
    \caption{Training parameters for counterfactual generation approaches for explaining policy $\pi$ in CitiBikes environment.}
    \begin{adjustbox}{width=\linewidth}
    \begin{tabular}{ccc} \toprule
    Algorithm                                                                               & Training Parameter           & Value \\ \midrule
    \multirow{6}{*}{GANterfactual-RL}                                                       & Dataset Size                 &     $5 \cdot 10^5$  \\
                                                                                            & Generator architecture       & [128, 128, 128]      \\
                                                                                            & Discriminator architecture   & [256, 256]    \\
                                                                                            & Learning rate (generator)    &  $10^{-4}$     \\
                                                                                            & Learning rate(discriminator) &  $10^{-4}$      \\
                                                                                            & Training timesteps           & 1000     \\ \midrule
    \multirow{1}{*}{RACCER-HTS}                                                             & Number of iterations         &  300     \\
                                                                                             \midrule
    \multirow{2}{*}{\begin{tabular}[c]{@{}l@{}}RACCER-Advance\\ RACCER-Rewind\end{tabular}} & Number of generations        &  24     \\
                                                                                            & Population size              &      100 \\ \bottomrule
    \end{tabular}
    \end{adjustbox}
    \label{table:TrainingParamsMethods}
\end{table}

\subsection{Results}
\label{SectionCitiBikes:Results}

Table \ref{table:results} shows the results of evaluation of counterfactual generation methods on the CitiBikes environment. 

\begin{table*}[t]
    \caption{Evaluation results of GANterfactual-RL, RACCER, RACCER-Advance, and RACCER-Rewind counterfactual generation methods for RL in CitiBikes task.}
    \begin{adjustbox}{width=\linewidth}
        \begin{tabular}{c||c||c|c|c|c} \toprule
                                                     &                           & \textbf{GANterfactual-RL} & \textbf{RACCER} & \textbf{RACCER-Advance} & \textbf{RACCER-Rewind} \\ \midrule
            \multirow{2}{*}{\textbf{Scalability}}        & Coverage ($\%$) ($\uparrow$) & 99.0 & 99.0 & 99.0 & 99.0 \\
                                                         & Pre-Training time (s) ($\downarrow$) & $5 \cdot 10^4$ & \textbf{0} & \textbf{0} & \textbf{0} \\
                                                         & Generation Time (s) ($\downarrow$) & \textbf{0.0.00073} & 191.2 & 34.82 & 70.66 \\ \midrule
            \multirow{2}{*}{\textbf{Feature Similarity}} & Proximity ($\uparrow$) & 0.5671 & 0.5567 & 0.5666 & 0.5596\\
                                                         & Sparsity ($\downarrow$) & 0.6879 & 0.3113 & 0.57 & 0.5934\\ \midrule
            \textbf{Plausibility ($\%$) ($\uparrow$)} &  & 82.0 & \textbf{100.0} & \textbf{100.0}  & \textbf{100.0} \\ \midrule
            \multirow{2}{*}{\textbf{Diversity}}          & Number of explanations ($\uparrow$) & 1.0 & 1.0 & 1.0 & 1.0\\
                                                         & Feature diversity ($\uparrow$) & 1.0 & 0.0 & 0.0 & 5.196 \\ \bottomrule                       
        \end{tabular}
    \end{adjustbox}
    \label{table:results}
\end{table*}

\subsubsection{Scalability}
\label{SectionCitiBikes:ResultsScalability}

All approaches achieve nearly perfect coverage, with a $99\%$ success rate in generating a counterfactual for each factual query. In terms of time efficiency, there is a clear difference between dataset-based and RL-based methods. The dataset-based GANterfactual-RL approach requires substantial pre-training time (around 16h). However, once trained, it can generate counterfactuals extremely quickly, in under a millisecond. Conversely, the RL-based methods require no pre-training, but the time needed to generate each counterfactual is significantly higher as they must traverse the agent's execution path. Among the RL-based approaches, RACCER-Advance is the fastest, followed by RACCER-Rewind and then RACCER. RACCER-Advance improves scalability considerably, reducing the generation time to less than 40 seconds.

Scalability is an important factor when developing counterfactual explanations for real-life tasks, as users require timely responses from the RL system—especially when explaining questionable or surprising decisions. When choosing a specific counterfactual method to explain decisions of a real-life self-adaptive system based on RL, developers need to consider the trade-off between pre-training and generation time. 

\subsubsection{Feature-based Similarity Metrics}
\label{SectionCitiBikes:ResultsSimilarity}

All approaches demonstrate comparable results on the proximity metric, with RACCER performing slightly better than others. On the sparsity metric, RACCER achieves the best result, followed by RACCER-Advance and RACCER-Rewind. GANterfactual-RL again performs the worst, with a sparsity of $0.6879$, indicating that, on average, a counterfactual and the original instance differ in nearly $68\%$ of their features.

\subsubsection{Plausibility}
\label{SectionCitiBikes:ResultsPlausibility}

We find that RL-based approaches RACCER, RACCER-Rewind, and RACCER-Advance produce only plausible instances. This is expected, as they search for the counterfactual among the environment's state space. On the other hand, the GANterfactual-RL approach only produces plausible counterfactuals in $82\%$ of cases. This is due to the generative nature of the approach, where a counterfactual state is generated without considering the potential causal relationships between features. It is still possible to use Ganterfactual, but it shows that careful filtering of generated counterfactuals is necessary. 

\subsubsection{Diversity}
\label{SectionCitiBikes:ResultsDiversity}

All approaches generate only one counterfactual per factual state. RACCER-Rewind shows a the higher feature diversity compared to the other algorithms, which have diversity scores close to zero. This suggests that RACCER-Rewind explores a wider range of alternative scenarios, whereas the other methods tend to produce very similar explanations.

\section{Conclusions and Future Work}

In this work, we present a Python library for generating counterfactual explanations, outline its key features, and evaluate its performance in a real-world CitiBikes scenario. We apply four counterfactual generation methods on a CitiBikes environment and compare them based on four evaluation metrics. We find that dataset-based methods quickly produce counterfactuals but require extensive pre-training. On the other hand, RL-based methods require substantially more time to generate counterfactuals, but produce more plausible counterfactuals that are more similar in features to the original state. 

We identified plausibility as the main challenge in using dataset-based counterfactual generation methods, while RL-based methods struggle with scalability in real-world domains. Nevertheless, this library provides the first implementation of both approaches within a unified framework, enabling their use in smaller-scale scenarios and controlled environments. This represents a valuable step toward helping users understand model behavior through counterfactual explanations.  

Future work should focus on incorporating plausibility constraints into dataset-based methods and exploring alternative optimization strategies and state space reduction techniques to improve the scalability of RL approaches. A limitation of current work is that the evaluation is restricted to quantitative metrics but does not measure how useful the explanations are to end users. Future work should then include user studies that help assess the usefulness of counterfactual explanations to support real-world decision-making tasks.

\begin{table}[t]
\label{TableOverviewMethods}
\caption{Overview of algorithms in EARL for generating counterfactual explanations for RL, categorized by search algorithms and requirements.}
\centering
    \begin{adjustbox}{width=\linewidth}
        \begin{tabular}{ccc} \toprule
        Algorithm  & Counterfactual Search & Requirements Type \\  \midrule
        GANterfactual-RL  & Generative Modelling &  Dataset-based \\
        RACCER-HTS    & Heuristic Tree Search   & RL-based \\
        RACCER-Advance  & NSGA-II  & RL-based \\
        RACCER-Rewind  & NSGA-II & RL-based \\ \bottomrule   
        \end{tabular}
    \end{adjustbox}

\end{table}

\begin{credits}
\subsubsection{\ackname} This publication has been supported in part by the Science Foundation Ireland under Grant number 18/CRT/6223 and Frontiers for the Future Grant No. 21/FFP-A/8957. For the purpose of Open Access, the author has applied a CC BY public copyright license to any Author Accepted Manuscript version arising from this submission.

\subsubsection{\discintname}
%It is now necessary to declare any competing interests or to specifically state that the authors have no competing interests. Please place the statement with a bold run-in heading in small font size beneath the (optional) acknowledgments\footnote{If EquinOCS, our proceedings submission system, is used, then the disclaimer can be provided directly in the system.}, for example: 
The authors have no competing interests to declare that are relevant to the content of this article. 
%Or: 
%Author A has received research grants from Company W. Author B has received a speaker honorarium from Company X and owns stock in Company Y. Author C is a member of committee Z.
\end{credits}
\bibliographystyle{splncs04}
\bibliography{ref}

@article{macias2013self,
  title={Self-adaptive systems: A survey of current approaches, research challenges and applications},
  author={Mac{\'\i}as-Escriv{\'a}, Frank D and Haber, Rodolfo and Del Toro, Raul and Hernandez, Vicente},
  journal={Expert Systems with Applications},
  volume={40},
  number={18},
  pages={7267--7279},
  year={2013},
  publisher={Elsevier}
}

@article{gajcin2024redefining,
  title={Redefining Counterfactual Explanations for Reinforcement Learning: Overview, Challenges and Opportunities},
  author={Gajcin, Jasmina and Dusparic, Ivana},
  journal={ACM Computing Surveys},
  volume={56},
  number={9},
  pages={1--33},
  year={2024},
  publisher={ACM New York, NY}
}

@inproceedings{van2024towards,
  title={Towards Understanding Trust in Self-adaptive Systems},
  author={Van Landuyt, Dimitri and Hal{\'a}sz, D{\'a}vid and Verreydt, Stef and Weyns, Danny},
  booktitle={Proceedings of the 19th International Symposium on Software Engineering for Adaptive and Self-Managing Systems},
  pages={207--213},
  year={2024}
}

@article{miller2019explanation,
  title={Explanation in artificial intelligence: Insights from the social sciences},
  author={Miller, Tim},
  journal={Artificial intelligence},
  volume={267},
  pages={1--38},
  year={2019},
  publisher={Elsevier}
}

@article{puri2019explain,
  title={Explain your move: Understanding agent actions using specific and relevant feature attribution},
  author={Puri, Nikaash and Verma, Sukriti and Gupta, Piyush and Kayastha, Dhruv and Deshmukh, Shripad and Krishnamurthy, Balaji and Singh, Sameer},
  journal={arXiv preprint arXiv:1912.12191},
  year={2019}
}

@article{verma2021counterfactual,
  title={Counterfactual Explanations for Machine Learning: Challenges Revisited},
  author={Verma, Sahil and Dickerson, John and Hines, Keegan},
  journal={arXiv preprint arXiv:2106.07756},
  year={2021}
}

@book{Sutton1998,
  author = {Sutton, Richard S. and Barto, Andrew G.},
  edition = {Second},
  publisher = {The MIT Press},
  title = {Reinforcement Learning: An Introduction},
  url = {http://incompleteideas.net/book/the-book-2nd.html},
  year = {2018 }
}

@article{saputri2020application,
  title={The application of machine learning in self-adaptive systems: A systematic literature review},
  author={Saputri, Theresia Ratih Dewi and Lee, Seok-Won},
  journal={IEEE Access},
  volume={8},
  pages={205948--205967},
  year={2020},
  publisher={IEEE}
}

@article{Gajcin2023RACCERTR,
  title={RACCER: Towards Reachable and Certain Counterfactual Explanations for Reinforcement Learning},
  author={Jasmina Gajcin and Ivana Dusparic},
  journal={ArXiv},
  year={2023},
  volume={abs/2303.04475},
  url={https://api.semanticscholar.org/CorpusID:257405415}
}

@article{wachter2017counterfactual,
  title={Counterfactual explanations without opening the black box: Automated decisions and the GDPR},
  author={Wachter, Sandra and Mittelstadt, Brent and Russell, Chris},
  journal={Harv. JL \& Tech.},
  volume={31},
  pages={841},
  year={2017},
  publisher={HeinOnline}
}

@inproceedings{li2020explanations,
  title={Explanations for human-on-the-loop: A probabilistic model checking approach},
  author={Li, Nianyu and Adepu, Sridhar and Kang, Eunsuk and Garlan, David},
  booktitle={Proceedings of the IEEE/ACM 15th International Symposium on Software Engineering for Adaptive and Self-Managing Systems},
  pages={181--187},
  year={2020}
}

@inproceedings{yigitbas2021enhancing,
  title={Enhancing human-in-the-loop adaptive systems through digital twins and VR interfaces},
  author={Yigitbas, Enes and Karakaya, Kadiray and Jovanovikj, Ivan and Engels, Gregor},
  booktitle={2021 International Symposium on Software Engineering for Adaptive and Self-Managing Systems (SEAMS)},
  pages={30--40},
  year={2021},
  organization={IEEE}
}

@article{olson2019counterfactual,
  title={Counterfactual states for atari agents via generative deep learning},
  author={Olson, Matthew L and Neal, Lawrence and Li, Fuxin and Wong, Weng-Keen},
  journal={arXiv preprint arXiv:1909.12969},
  year={2019}
}

@article{huber2023ganterfactual,
  title={Ganterfactual-rl: Understanding reinforcement learning agents' strategies through visual counterfactual explanations},
  author={Huber, Tobias and Demmler, Maximilian and Mertes, Silvan and Olson, Matthew L and Andr{\'e}, Elisabeth},
  journal={arXiv preprint arXiv:2302.12689},
  year={2023}
}

@inproceedings{greydanus2018visualizing,
  title={Visualizing and understanding atari agents},
  author={Greydanus, Samuel and Koul, Anurag and Dodge, Jonathan and Fern, Alan},
  booktitle={International Conference on Machine Learning},
  pages={1792--1801},
  year={2018},
  organization={PMLR}
}

@article{skirzynski2021automatic,
  title={Automatic discovery of interpretable planning strategies},
  author={Skirzy{\'n}ski, Julian and Becker, Frederic and Lieder, Falk},
  journal={Machine Learning},
  volume={110},
  pages={2641--2683},
  year={2021},
  publisher={Springer}
}

@inproceedings{amir2018highlights,
  title={Highlights: Summarizing agent behavior to people},
  author={Amir, Dan and Amir, Ofra},
  booktitle={Proceedings of the 17th international conference on autonomous agents and multiagent systems},
  pages={1168--1176},
  year={2018}
}

@inproceedings{sovrano2018making,
  title={Making things explainable vs explaining: Requirements and challenges under the GDPR},
  author={Sovrano, Francesco and Vitali, Fabio and Palmirani, Monica},
  booktitle={International Workshop on AI Approaches to the Complexity of Legal Systems},
  pages={169--182},
  year={2018},
  organization={Springer}
}

@article{metzger2023user,
  title={A user study on explainable online reinforcement learning for adaptive systems},
  author={Metzger, Andreas and Laufer, Jan and Feit, Felix and Pohl, Klaus},
  journal={ACM Transactions on Autonomous and Adaptive Systems},
  year={2023},
  publisher={ACM New York, NY}
}

@inproceedings{gajcin2022reccover,
  title={Reccover: Detecting causal confusion for explainable reinforcement learning},
  author={Gajcin, Jasmina and Dusparic, Ivana},
  booktitle={International Workshop on Explainable, Transparent Autonomous Agents and Multi-Agent Systems},
  pages={38--56},
  year={2022},
  organization={Springer}
}

@misc{MARO_MSRA,
title = {MARO: A Multi-Agent Resource Optimization Platform},
author = {Jiang, Arthur and Zhang,  Jia and Yu, Pingchao and Huang, Lyuchun and Qiu, Yang and Wang, Jinyu and Shi, Wenlei and Li, Kaiqi and Wang, Zhanyu and Zhang, Chengruidong and Sun, Tianyi and Chen, Miaoran and Yu, Kuanwei and Wei, Xinran and Li, Michael and Shang, Ning and Meng, Qiwei and Li, Shan and Bian, Jiang and Cheng, Biao and Liu, Tie-Yan},
year = {2020},
publisher = {GitHub},
journal = {GitHub repository},
url = {https://github.com/microsoft/maro}
}

@article{milani2024explainable,
  title={Explainable reinforcement learning: A survey and comparative review},
  author={Milani, Stephanie and Topin, Nicholay and Veloso, Manuela and Fang, Fei},
  journal={ACM Computing Surveys},
  volume={56},
  number={7},
  pages={1--36},
  year={2024},
  publisher={ACM New York, NY}
}

@inproceedings{byrne2019counterfactuals,
  title={Counterfactuals in explainable artificial intelligence (XAI): Evidence from human reasoning.},
  author={Byrne, Ruth MJ},
  booktitle={IJCAI},
  pages={6276--6282},
  year={2019},
  organization={California, CA}
}

@article{cadario2021understanding,
  title={Understanding, explaining, and utilizing medical artificial intelligence},
  author={Cadario, Romain and Longoni, Chiara and Morewedge, Carey K},
  journal={Nature human behaviour},
  volume={5},
  number={12},
  pages={1636--1642},
  year={2021},
  publisher={Nature Publishing Group UK London}
}

@inproceedings{d2019learning,
  title={On learning in collective self-adaptive systems: State of practice and a 3d framework},
  author={D'Angelo, Mirko and Gerasimou, Simos and Ghahremani, Sona and Grohmann, Johannes and Nunes, Ingrid and Pournaras, Evangelos and Tomforde, Sven},
  booktitle={2019 IEEE/ACM 14th International Symposium on Software Engineering for Adaptive and Self-Managing Systems (SEAMS)},
  pages={13--24},
  year={2019},
  organization={IEEE}
}

@article{schulman2017proximal,
  title={Proximal policy optimization algorithms},
  author={Schulman, John and Wolski, Filip and Dhariwal, Prafulla and Radford, Alec and Klimov, Oleg},
  journal={arXiv preprint arXiv:1707.06347},
  year={2017}
}

@INPROCEEDINGS{heimerson2022adaptiveCooling,
  author={Heimerson, Albin and Sjölund, Johannes and Brännvall, Rickard and Gustafsson, Jonas and Eker, Johan},
  booktitle={2022 IEEE International Conference on Autonomic Computing and Self-Organizing Systems Companion (ACSOS-C)}, 
  title={Adaptive Control of Data Center Cooling using Deep Reinforcement Learning}, 
  year={2022}, 
  pages={1-6}, 
  doi={10.1109/ACSOSC56246.2022.00018}
}

@INPROCEEDINGS{rosero2024DwnEws,
  author={Rosero, Juan C. and Cardozo, Nicolás and Dusparic, Ivana},
  booktitle={2024 IEEE International Conference on Autonomic Computing and Self-Organizing Systems Companion (ACSOS-C)}, 
  title={Multi-Objective Deep Reinforcement Learning Optimisation in Autonomous Systems}, 
  year={2024}, 
  pages={97-102}, 
  doi={10.1109/ACSOS-C63493.2024.00038}}

@INPROCEEDINGS{hossain2023covernav,
  author={Hossain, Jumman and Faridee, Abu-Zaher and Roy, Nirmalya and Basak, Anjan and Asher, Derrik E.},
  booktitle={2023 IEEE International Conference on Autonomic Computing and Self-Organizing Systems (ACSOS)}, 
  title={CoverNav: Cover Following Navigation Planning in Unstructured Outdoor Environment with Deep Reinforcement Learning}, 
  year={2023}, 
  pages={127-132}, 
  doi={10.1109/ACSOS58161.2023.00030}}

@article{gajcin2023raccer,
  title={Raccer: Towards reachable and certain counterfactual explanations for reinforcement learning},
  author={Gajcin, Jasmina and Dusparic, Ivana},
  journal={23rd International Conference on Autonomous Agents and Multiagent Systems
(AAMAS 2024)},
  year={2023}
}

\end{document}